\documentclass{article}

\usepackage[preprint,nonatbib]{neurips_2025}

\usepackage[utf8]{inputenc} 
\usepackage[T1]{fontenc}    
\usepackage{hyperref}       
\usepackage{url}            
\usepackage{booktabs}       
\usepackage{amsfonts}       
\usepackage{nicefrac}       
\usepackage{microtype}      
\usepackage{xcolor}         
\usepackage{graphicx}
\usepackage{subcaption}
\usepackage{float}
\usepackage{booktabs}
\usepackage{wrapfig}
\usepackage{tabularx}
\usepackage{makecell}
\usepackage{caption}
\usepackage{amsmath}

\usepackage{amssymb}
\usepackage[noend]{algpseudocode} 
\usepackage{algorithm}
\usepackage{algpseudocode}
\usepackage{enumitem}
\usepackage{biblatex}
\title{RDPO: Real Data Preference Optimization for Physics Consistency Video Generation}

\author{
Wenxu Qian$^{1,2}$\textsuperscript{*},
Chaoyue Wang$^{2}$\textsuperscript{*},
Hou Peng$^{2}$,
Zhiyu Tan$^{1}$,
Hao Li$^{1}$\textsuperscript{\textdagger},
Anxiang Zeng$^{2}$\textsuperscript{\textdagger} \\
$^{1}$Fudan University \quad
$^{2}$Shopee Inc.
}

\begin{document}

\makeatletter
\def\blfootnote{\gdef\@thefnmark{}\@footnotetext} 
\makeatother

\blfootnote{* Equal contribution.} 
\blfootnote{\textdagger Corresponding author.}
\blfootnote{This work was completed during an internship at Shopee.}

\maketitle

\begin{abstract}

Video generation techniques have achieved remarkable advancements in visual quality, yet faithfully reproducing real‑world physics remains elusive. Preference‑based model post-training may improve physical consistency, but requires costly human‑annotated datasets or reward models that are not yet feasible. To address these challenges, we present Real Data Preference Optimisation (RDPO), an annotation‑free framework that distils physical priors directly from real‑world videos. Specifically, the proposed RDPO reverse‑samples real video sequences with a pre‑trained generator to automatically build preference pairs that are statistically distinguishable in terms of physical correctness. A multi‑stage iterative training schedule then guides the generator to obey physical laws increasingly well. Benefiting from the dynamic information explored from real videos, our proposed RDPO significantly improves the action coherence and physical realism of the generated videos. Evaluations on multiple benchmarks and human evaluations have demonstrated that RDPO achieves improvements across multiple dimensions. The source code and demonstration of this paper are available in \href{https://wwenxu.github.io/RDPO/}{webpage} .

\end{abstract}

\section{Introduction}
\label{sec:intro}

The field of video generation is undergoing rapid development \cite{xing2024survey,}, driven by its immense application potential in diverse domains such as entertainment, education, and advanced simulations. A primary goal in this field is to create videos that are high-quality, temporal coherence, and high-fidelity. Beyond mere visual appeal, a key frontier is achieving \textit{physical consistency}, ensuring that the dynamics and interactions depicted in generated videos adhere to real-world laws \cite{lin2025exploring}.

Prior work \cite{hong2022cogvideo,guo2023animatediff,guo2024i2v,wu2024fairy,kong2024hunyuanvideo} has shown that large‑scale video datasets enable models to learn certain physical regularities Nevertheless, existing video generators still struggle to reproduce complex physical processes faithfully. Although pre‑training exposes a model to a vast spectrum of motion patterns, many of the learned dynamics remain only superficially plausible. Consider, for example, pouring water into a cup: while a model may capture the intuitive rise of the water level, it often fails to maintain the correct quantitative relationship between the poured volume and the height increment. Simply minimizing the video reconstruction (or noise‑prediction) loss does not guarantee better physical consistency, making the acquisition of reliable physical priors particularly challenging.

Recently, works~\cites{prabhudesai2024video,wu2024boosting,Yuan_2024_CVPR,lee2023aligningtexttoimagemodelsusing,black2024trainingdiffusionmodelsreinforcement,yang2025ipoiterativepreferenceoptimization,xu2025visionrewardfinegrainedmultidimensionalhuman,liu2025improvingvideogenerationhuman,diffusion_dpo,li2024aligning,} found that post-training techniques, particularly preference optimization methods, could take a further step to help the video generation model avoid generating incorrect dynamics and enhance physics consistency. However, existing approaches depend on either costly human annotations or highly accurate reward models. Building a reward function that detects physical violations in arbitrary videos remains an open problem. Moreover, assembling large‑scale human preference datasets is expensive, time‑consuming, and especially onerous when the criterion is physics rather than perceptual quality. Human judgments also vary, undermining the objectivity needed for robust learning. These limitations reveal a pressing need for an efficient, annotation‑free preference optimization strategy that explicitly targets physical realism.

\begin{figure*}[t]
    \centering
    \includegraphics[width=\textwidth]{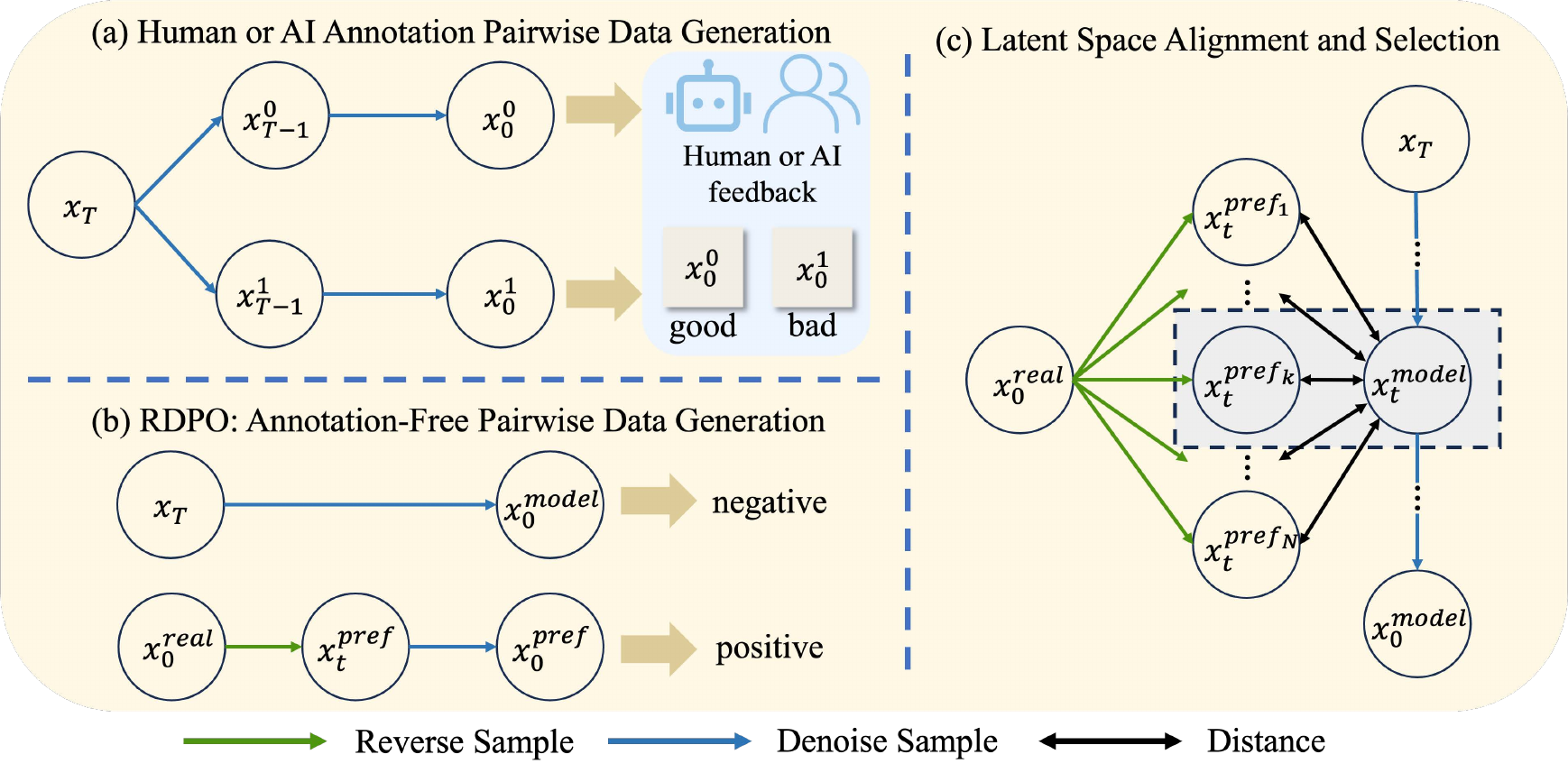} 
    \caption{
    Comparison of (a) Human or AI Annotation Pairwise Data Generation and (b) RDPO: Annotation-Free Pairwise Data Generation. (c) Latent Space Alignment and Selection. RDPO utilizes real data to automatically generate pairwise data.
    }
    \label{fig:intro_algorithm}
\end{figure*}
To overcome these limitations, we propose an annotation-free preference optimization framework named \textit{Real Data Preference Optimization (RDPO)}. The core motivation is to automatically construct preference training data using a pre-trained video generation model and real-world data. We observed that, accompanied by reducing the steps of reverse sampling,  the generated videos present a nearly continuous transition from the generated distribution to real-world physics consistency. 
Within these reverse sampling steps, RDPO constructs videos by selectively utilizing latent representations. These latent representations, derived from the reverse sampling process applied to real-world videos, possess two crucial characteristics. Firstly, they are rich in authentic physical information precisely because they originate from real-world data. Secondly, they are carefully chosen or guided to remain aligned with the model's inherent generative distribution; this alignment is vital for ensuring that the introduction of new physical priors does not drastically alter the model's learned visual style or lead to out-of-distribution outputs, thereby maintaining overall video quality. 
This method ensures that the generated videos possess real-world physical consistency. At the same time, it keeps the distributional deviation from becoming too large. 

Statistically, positive and negative preference data pairs can be built for a specific video generation model as long as it cannot generate physics consistency perfectly. Overall, the proposed RDPO offers several key advantages: it is \textbf{annotation-free}, contrasting sharply with traditional DPO or RLHF methods that depend on human input. By \textbf{leveraging real data}, RDPO treats authentic video footage as the "gold standard" for physical dynamic information. Finally, RDPO explores a new efficient pathway to leverage and learn the physical priors inherent in real-world video, instead of training on those videos again and again. The main contributions of this paper are as follows:

\begin{itemize}
    \item We propose Real Data Preference Optimization (RDPO), a novel, annotation-free preference optimization framework that is capable of improving the physical realism of generated videos.
    \item We demonstrate an effective method for automatically constructing preference pairs using the current video generation model and real-world videos, enabling the model to learn underlying physical laws.
    \item We validate the effectiveness of RDPO in improving the physics consistency and overall video quality of different baseline models through extensive experiments.
    \item We conduct a comparative analysis of RDPO against traditional DPO methods that rely on manual annotation and explore the potential synergistic benefits of combining both approaches.
\end{itemize}

\section{Related Work}
\label{related_work}

\paragraph{Video Generation.}
The field of video generation has recently seen significant advancements, largely driven by the rapid development of diffusion models \cite{ho2020denoisingdiffusionprobabilisticmodels,Sohl-Dickstein_Weiss_Maheswaranathan_Ganguli_2015,Song_Meng_Ermon_2020,Zhang_Tao_Chen_2022}. These models have greatly enhanced the diversity, realism, and overall quality of generated video content \cite{Ho_Chan_Saharia_Whang_Gao_Gritsenko_Kingma_Poole_Norouzi_Fleet_et,Singer_Polyak_Hayes_Yin_An_Zhang_Hu_Yang_Ashual_Gafni_et,Wu_Huang_Zhang_Li_Ji_Yang_Sapiro_Duan_2021}.
Transformer-based architectures, such as DiT and U-ViT \cite{bao2023all,peebles2023scalable}, have augmented or replaced traditional U-Net frameworks, significantly improving spatio-temporal coherence and model scalability \cite{blattmann2023alignlatentshighresolutionvideo,wang2023modelscopetexttovideotechnicalreport,guo2024animatediffanimatepersonalizedtexttoimage}.
In addition to these architectural advancements, the continuous expansion of model parameters and training data has propelled progress in the field \cite{Ho_Chan_Saharia_Whang_Gao_Gritsenko_Kingma_Poole_Norouzi_Fleet_et,Singer_Polyak_Hayes_Yin_An_Zhang_Hu_Yang_Ashual_Gafni_et,Wu_Huang_Zhang_Li_Ji_Yang_Sapiro_Duan_2021}. 
Despite these remarkable achievements, current models still face challenges in ensuring physics consistency, maintaining logical consistency in complex interactions. Continued research and development are essential to addressing these challenges and further advancing the field.


\paragraph{Physical Optimization in Video Generation.}
Learning physical priors from real-world data is crucial for enhancing the physical plausibility of generated videos. While training on large-scale video datasets \cite{brooks2024video,bruce2024genie,yang2023learning} allows models to implicitly internalize some physics through exposure to diverse dynamic scenes, this understanding is often superficial and struggles with novel or complex scenarios. Thus, more direct methods, such as incorporating external feedback mechanisms, are explored \cite{furuta2024improving,yang2025ipo,liu2025improving}.

Among these, post-training fine-tuning has gained traction for its ability to steer powerful pre-trained models towards physical correctness. These strategies frequently employ reward models, trained on human preferences or AI-generated feedback signals, to score generated content \cite{furuta2024improving,yang2025ipo,liu2025improving,huang2024gen,xue2024phyt2v}. Drawing inspiration from RLHF paradigms, these methods iteratively refine the video generator to align outputs with desired physical behaviors and improve dynamic interactions, thereby addressing common issues like object interpenetration or defiance of gravity.

However, these approaches face substantial challenges. Learning complex physical laws accurately from high-dimensional observational data remains difficult \cite{brooks2024video,kang2024far}. Concurrently, creating robust and generalizable reward models to reliably detect diverse physical violations, without being easily exploited by the generator, continues to be an open research problem.

Human annotation for post-training is expensive and time-consuming, particularly for physics, which needs expert review and suffers from rater variability\cite{xu2025visionrewardfinegrainedmultidimensionalhuman}. While scalable, AI evaluators can have flawed physics knowledge, inherit biases, struggle with out-of-distribution data, or be exploited by generators to achieve high scores without truly improving physical realism.

\section{Method}
\label{sec:method}


\subsection{Preliminaries}

\paragraph{Rectified Flow.}
Rectified Flow \cite{reflow} establishes approximately straight, constant-velocity trajectories between a data distribution $\pi_0$ (samples $X_0$) and a prior $\pi_1$ (samples $X_1$). It learns a velocity field $v(x, t)$ for the ordinary differential equation (ODE) $\frac{dx_t}{dt} = v(x_t, t)$ (mapping $X_0$ at $t=0$ to $X_1$ at $t=1$). The velocity field $v$ is trained by minimizing the following loss function $\mathcal{L}(v)$:
\begin{equation} \label{eq:reflow_loss_function}
    \mathcal{L}(v) = \int_{0}^{1} E \left[ \left\| (X_1 - X_0) - v\left((1-t)X_0 + tX_1, t\right) \right\|^2 \right] dt
\end{equation}
This loss function encourages the learned velocity $v$ along the linear path $(1-t)X_0 + tX_1$ to align with the constant velocity $(X_1 - X_0)$ of a straight trajectory. For generation, a sample $x_1 \sim \pi_1$ is drawn, and the ODE is integrated backward from $t=1$ to $t=0$ to produce $x_0'$ approximating $\pi_0$.

\paragraph{Direct Preference Optimization.}
Direct Preference Optimization (DPO) \cite{diffusion_dpo} is a robust method for fine-tuning generative models, such as language models, to better align with human preferences. Unlike traditional reinforcement learning approaches, DPO leverages preference data to optimize the model's output by estimating a reward model implicitly. In this framework, preference pairs \((x_w, x_l)\) are used, where \(x_w\) is the preferred sample and \(x_l\) is the less preferred one. The optimization process aims to maximize the likelihood of generating \(x_w\) while minimizing the likelihood of generating \(x_l\), relative to a reference model \(\pi_{ref}\). This is achieved by directly optimizing the conditional distribution \(p_\theta(x_0|c)\), where \(x_0\) is the generated output given the condition \(c\). The objective function can be expressed as:

\begin{align}
L_{DPO}(\theta) = -\mathbb{E}_{c, x_w, x_l}\left[\log \sigma\left(\beta \log \frac{p_\theta(x_w|c)}{p_\text{ref}(x_w|c)} - \beta \log \frac{p_\theta(x_l|c)}{p_\text{ref}(x_l|c)}\right)\right]
\end{align}

\subsection{Real Data Preference Optimization (RDPO)}
\label{sec:rdpo}

Existing post-training paradigms for text-to-video diffusion models suffer from two core limitations:  
(i) accurately capturing the underlying physical laws that govern real-world dynamics, and  
(ii) the heavy cost of collecting human preference annotations.  
We propose \textit{Real Data Preference Optimization} (RDPO) to overcome both issues.  
RDPO automatically constructs preference data by combining \emph{random} sampling with \emph{reverse} (i.e., latent-guided) sampling, steering generation toward outputs that faithfully respect physical consistency. By eliminating manual annotation, RDPO dramatically reduces the expense of traditional preference-based fine-tuning.

\begin{figure*}[!htbp]
    \centering
    \includegraphics[width=\textwidth]{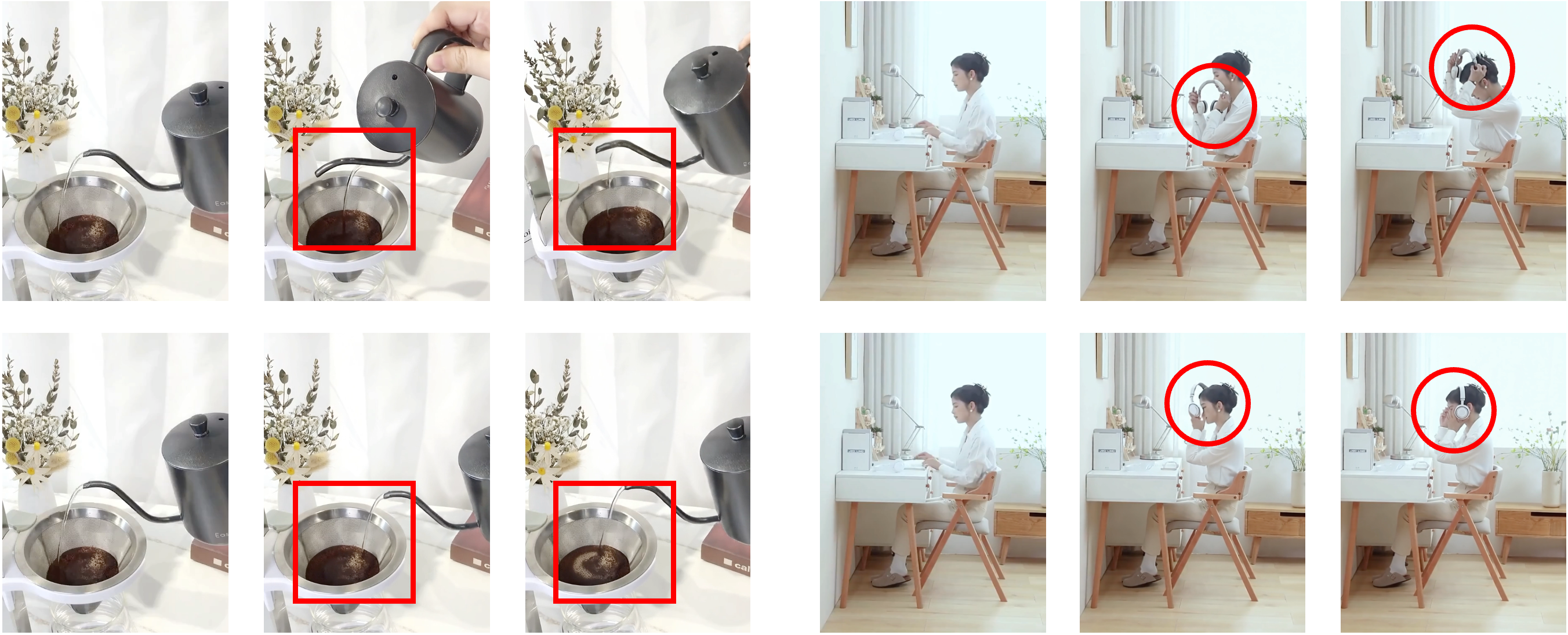}
    \caption{Visual comparison of video generation.  
    \textbf{Top}: Videos generated by denoising from pure noise.  
    \textbf{Bottom}: Videos generated by reverse sampling from an intermediate latent.  
    Reverse sampling produces substantially better physical consistency and overall fidelity.}
    \label{fig:sample}
\end{figure*}

Our method is inspired by image-to-image editing and style-transfer pipelines \cite{stablediffusion,meng2021sdedit}, which denoise a noisy real image rather than synthesizing solely from Gaussian noise.  
We extend this idea to video: starting from a real video, we inject noise at a user-controlled diffusion step and then \emph{reverse sample} to obtain a clean video.  
Figure~\ref{fig:sample} shows that reverse sampling, when used to guide generation, yields videos with markedly more plausible motion trajectories and natural object interactions than conventional denoise-from-noise sampling.  
Furthermore, we introduce \emph{rejection sampling} to select higher-quality latent pairs (see below).

\subsubsection{Annotation-Free Pairwise Data Generation}
\label{sec:pairwise_generation}
Gradually decreasing the starting step of reverse sampling controls the noise level added to the real video, giving a near-continuous interpolation from the model’s own distribution to one that closely matches real-world physics.  
Algorithm~\ref{alg:diffusion_preference_generation} details this fully automated construction pipeline.

Instead of naïvely reverse-sampling every noisy latent, we adopt rejection sampling.  
Given a real video’s noisy latent \(\hat{x}_t\), we select the instance that is closest—in latent space—to the model’s own sample \(x_t^{\text{model}}\) drawn at the same timestep.  
In practice, the Euclidean (L2) distance between latents provides a simple yet effective similarity metric.  
The subsequent denoising then proceeds from this selected \(\hat{x}_t\).  
This strategy injects real-world information while limiting divergence from the model’s native manifold, thus improving training stability.

The procedure yields two complementary video categories:  
\(\smash{x_{\text{preferred}}}\), obtained via reverse sampling and therefore enriched with physical priors, and  
\(\smash{x_{\text{model}}}\), obtained via standard denoise sampling.  
Empirically, \(x_{\text{preferred}}\) exhibits stronger physical consistency, so each pair \((x_{\text{preferred}}, x_{\text{model}})\) forms an automatically generated preference example.  
This preference is \emph{statistical}: for the vast majority of pairs, the first element displays superior physics.

\subsubsection{Preference Optimization Objective}
\label{sec:rdpo_loss}
Given the automatically generated preference set, we fine-tune the model with the Flow-DPO loss:
\begin{align}
\mathcal{L}_{\text{RDPO}}
  = -\mathbb{E}\Bigl[ \log \sigma\!\Bigl(
      \beta \Bigl(
        \log \tfrac{\pi_{\theta}(x_{\text{preferred}}\!\mid c)}
                   {\pi_{\text{ref}}(x_{\text{preferred}}\!\mid c)}
        - \log \tfrac{\pi_{\theta}(x_{\text{model}}\!\mid c)}
                     {\pi_{\text{ref}}(x_{\text{model}}\!\mid c)}
      \Bigr)
    \Bigr) \Bigr],
\end{align}
where \(\pi_{\text{ref}}\) is a fixed pretrained reference policy, \(\sigma(\cdot)\) is the sigmoid, and \(c\) denotes the text prompt.  
Minimizing \(\mathcal{L}_{\text{RDPO}}\) raises the likelihood of \(x_{\text{preferred}}\) relative to \(x_{\text{model}}\), implicitly teaching the model to respect physical laws and natural motion.

\subsubsection{Progressive Training Strategy}
\label{sec:progressive_training}
To absorb real-world dynamics both stably and effectively, we adopt a \emph{coarse-to-fine} curriculum over reverse-sampling steps.  
{Phase~1}: We start from a \emph{large} diffusion step (heavy noise), encouraging the model to learn macroscopic motion patterns and global structure.  
{Phase~2}: Once the generated distribution approaches real videos—i.e., glaring artifacts largely disappear—we progressively reduce the noise level.  
Lower-noise stages focus on finer physical interactions and texture details.  
Crucially, each later phase continues to interleave data synthesized in previous phases, ensuring that newly learned details do not overwrite earlier, large-scale dynamics.  
Section~\ref{secc:impact_of_progress} analyzes the influence of different starting steps.

\begin{minipage}{0.48\textwidth}
\begin{algorithm}[H]
\caption{RDPO: Annotation-Free Pairwise Data Generation}
\label{alg:diffusion_preference_generation}
\begin{algorithmic}[1]
\Require
    Real video dataset $D_{\text{real}}$;
    Video model denoise $\pi_{\theta}(\cdot, \cdot)$;
    Video model forward $f(\cdot, \cdot)$;
    Total sample steps $T$;
    Reverse sample step $t$;
    Sampling times $K$.
\Ensure
    Pairwise data $(x_0^{\text{preferred}}, x_0^{\text{model}})$.

\State Sample $x_0^{\text{real}} \sim D_{\text{real}}$.
\State $\hat{x}_T \leftarrow f(x_0^{\text{real}}, T)$.
\State $x_{T-1}^{\text{model}}, \ldots, x_1^{\text{model}}, x_0^{\text{model}} \leftarrow \pi_{\theta}(\hat{x}_T, T)$.

\State $\text{min\_distance} \leftarrow \infty$.
\For{$k = 1$ \textbf{to} $K$}
    \State $x_t \leftarrow f(x_0^{\text{real}}, t)$.
    \State $d \leftarrow \|x_t - x_t^{\text{model}} \|^2$.
    \If{$d < \text{min\_distance}$}
        \State $\ x_t^\textbf{preferred} \leftarrow x_t$.
        \State $\text{min\_distance} \leftarrow d$.
    \EndIf
\EndFor

\State $x_0^{\text{preferred}} \leftarrow \pi_{\theta}(\hat{x}_t, t)$.
\State \Return $(x_0^{\text{preferred}}, x_0^{\text{model}})$.
\end{algorithmic}
\end{algorithm}
\end{minipage}
\hfill
\begin{minipage}{0.48\textwidth}
\begin{algorithm}[H]
\caption{RDPO: Progressive Training}
\label{alg:rdpo_training}
\begin{algorithmic}[1]
\Require
    Pairwise dataset $D_{\text{pairwise}}$;
    Learning rate $ \eta$;
    Real video dataset $D_{\text{real}}$;
    Initial model $\theta$;
    Reference model $\theta_{ref}$;
\Ensure
    Optimized model parameters $\theta$.
\State Initialize: $\theta_{ref} \leftarrow \theta$
\State $\text{step} \leftarrow 0$
\Repeat
    \State $\text{step} \leftarrow \text{step} + 1$
    \If{step \% 2 = 0}
        \State $(x_0^{\text{preferred}}, x_0^{\text{model}}) \sim D_{\text{pairwise}}$
        \State $\mathcal{L}_\theta \leftarrow \mathcal{L}_{\text{RDPO}}(\theta, \theta_{ref})$
        \State $\theta \leftarrow \theta - \eta\nabla_\theta\mathcal{L}_{\theta}$
    \Else
        \State $x_{\text{real}} \sim D_{\text{real}}$
        \State $\mathcal{L}_{\theta} \leftarrow \mathcal{L}_{\text{sft}}(\theta)$
        \State $\theta \leftarrow \theta - \eta\nabla_\theta\mathcal{L}_{\theta}$
    \EndIf
\Until{convergence}
\State \Return $\theta$
\vspace{0.5em} 
\end{algorithmic}
\end{algorithm}
\end{minipage}

\section{Experiments}
\label{sec:exp_result}

To validate the effectiveness of the proposed  Real Data Preference Optimization (RDPO), we adopt two video‑generation baselines and conduct a comprehensive series of experiments.  
First, we employ the open‑source model LTX‑Video(0.9.1)‑2B \cite{hacohen2024ltx}, a mid‑sized yet highly capable generator that offers efficient training and inference.  
In addition, we evaluate RDPO on a larger model, Shopee-MUG-V‑10B. 
Using Shopee-MUG-V‑10B allows us to probe RDPO’s capacity to enhance physical realism in complex daily‑life scenarios and human, object interactions.

Below, we present results for each baseline model in turn.

\subsection{Experiments on the \texorpdfstring{LTX‑2B}{LTX‑2B} Baseline}

\subsubsection{Implementation Details}

We train and evaluate using the open-source LTX-Video-2B model on publicly available data, ensuring full reproducibility. Training utilized the public {WISA} dataset~\cite{wang2025wisa}, an aggregation of diverse videos showcasing natural physical phenomena (e.g., fluid dynamics, deformable solids, particle systems) aimed at enhancing the model's understanding of physical laws.

RDPO post-training on the LTX-2B baseline was conducted on 32 NVIDIA H100 GPUs, employing the AdamW optimizer (learning rate $1\times10^{-5}$) with a per-GPU batch size of 8, resulting in a global batch size of 256. We applied LoRA with rank 256 and adopted the progressive training strategy featuring staged reverse-sampling steps, as detailed in Section~\ref{sec:method}. Specifically, the initial two RDPO stages used progressively shortened reverse-sampling schedules, while the third stage was trained on a balanced mix of data derived from the preceding stages. Each stage comprised 8k preference pairs for training.

For evaluation, we employed two public benchmarks: {VBench}~\cite{huang2024vbench++} for assessing overall video quality, and {Physics IQ}~\cite{motamed2025generative} for a fine-grained assessment of physical realism, covering domains like solid mechanics, fluids, and optics.
The public availability of the model, training data, and evaluation benchmarks ensures that our LTX-2B experiments are fully reproducible, thereby facilitating transparent and fair future comparisons.

\subsubsection{Effectiveness of RDPO Training}

\begin{table}[H]
  \caption{Isolate the contributions of RDPO and SFT Training}
  \centering
  \renewcommand\theadfont{\normalsize\bfseries}
  \resizebox{\textwidth}{!}{ 
  \begin{tabular}{l *{14}{c}}
    \toprule
    \textbf{Models} 
    & \makecell{Physics\\IQ} 
    & \makecell{Total\\Score} 
    & \makecell{I2V\\Score} 
    & \makecell{Qulity\\Score} 
    & \makecell{Subject\\Consistency} 
    & \makecell{Background\\Consistency} 
    & \makecell{Motion\\Smoothness} 
    & \makecell{Aesthetic\\Quality} 
    & \makecell{Imaging\\Quality} 
    & \makecell{I2V\\Subject} 
    & \makecell{I2V\\Background} 
    & \makecell{Camera\\Motion} \\
    \midrule
    LTX-2B  & 24.40 & 85.35 & 92.85 & 77.85 & 96.76 & 98.36 & 99.25  & 54.62 & 63.96 & 96.33 & 97.83 & 22.41 \\
    SFT  & 24.63 & 85.49 & 92.74 & 78.24 & 95.82 & 98.73 & 99.20 & 54.46 & 63.81 & 95.93 & 97.85 & 24.25 \\
    RDPO (w/o sft) & 24.71 & 85.54 & \textbf{93.00} & 78.08 & 96.52 & 98.57 & \textbf{99.30}  & 54.89 & 62.96 & 96.29 & 97.82 & \textbf{26.00} \\
    RDPO (w. sft) & \textbf{25.21} & \textbf{85.63} & 92.98 & \textbf{78.27} & \textbf{97.04} & \textbf{98.81} & 99.27  & \textbf{55.01} & \textbf{65.11} & \textbf{96.39} & \textbf{97.87} & 23.72 \\
    
    \bottomrule
  \end{tabular}
  }
  \label{tab:sft_effect}
\end{table}

To isolate the contribution of the Real Data Preference Optimization (RDPO) strategy, we compare different training approaches: (i) the baseline model, (ii) using Supervised Fine-tuning (SFT) alone, (iii) applying the RDPO strategy alone, and (iv) combining the RDPO strategy with SFT. Results are summarized in Table~\ref{tab:sft_effect}. 
As indicated by the results, applying the RDPO strategy alone brings improvements in metrics related to physical realism (Physics IQ) and enhances most aspects of overall video quality compared to the baseline and standard SFT methods. The approach that combines the RDPO strategy with SFT, as demonstrated by the RDPO (w. sft) model trained with real data, further contributes to training stability and yields the most favorable performance across key indicators, including physical realism and overall video quality.

Based on the quantitative comparisons, we observe that the RDPO strategy is particularly effective in significantly improving motion smoothness and aspects of physical accuracy, while standard SFT tends to benefit visual quality attributes. Ultimately, combining SFT and the RDPO strategy effectively integrates their respective strengths, leading to superior and comprehensive overall results. Visual comparisons are provided in the supplementary material.

\subsubsection{RDPO v.s.\ Human‑Annotated DPO}

In this section, we compare our proposed annotation-free RDPO approach with Direct Preference Optimization (DPO) trained on human-annotated data, and also examine their combination. We constructed 8,000 RDPO preference pairs from real data. For comparison, we employed expert annotators to create 1,000 human-labeled pairs—a process that was notably slower, roughly ten times the generation time of the RDPO data.

As presented in Table~\ref{tab:human_feedback}, RDPO (Annotation-Free) achieves performance that surpasses human-annotated DPO on several key metrics, including Physics IQ and Total Score, despite its significantly lower data acquisition cost. Furthermore, the hybrid model that merges RDPO and human preferences yields the strongest results across the majority of metrics in this comparison, clearly demonstrating that these two distinct sources of supervision are complementary and their combination is highly beneficial.

\begin{table}[H]
  \caption{Comparison of RDPO Annotation-Free Data and Human-label Data Training}
  \centering
  \renewcommand\theadfont{\normalsize\bfseries}
  \resizebox{\textwidth}{!}{ 
  \begin{tabular}{l *{13}{c}}
    \toprule
    \textbf{Models} 
    & \makecell{Physics\\IQ} 
    & \makecell{Total\\Score} 
    & \makecell{I2V\\Score} 
    & \makecell{Qulity\\Score} 
    & \makecell{Subject\\Consistency} 
    & \makecell{Background\\Consistency} 
    & \makecell{Motion\\Smoothness} 
    & \makecell{Aesthetic\\Quality} 
    & \makecell{Imaging\\Quality} 
    & \makecell{I2V\\Subject} 
    & \makecell{I2V\\Background} 
    & \makecell{Camera\\Motion} \\
    \midrule
    LTX-2B  & 24.40  & 85.35 & 92.85 & 77.85 & 96.76 & 98.36 & 99.25  & 54.62 & 63.96 & 96.33 & 97.83 & 22.41 \\
    Human-label DPO &  24.70 & 85.44 & 93.06 & 77.82 & 96.83 & 98.88 & 99.29 &  54.96 & 64.21 & 96.50 & 97.89 & 23.94 \\
    RDPO (Annotation-Free)& 25.21 & 85.63 & 92.98 & \textbf{78.27} & \textbf{97.04} & 98.81 & 99.27  &\textbf{ 55.01} & \textbf{65.11} & 96.39 & 97.87 & 23.72 \\
    RDPO (Mix Human-label) & 25.25 & \textbf{85.68} & \textbf{93.11} & 78.24 & 96.83 & \textbf{98.88} & \textbf{99.29} &  54.96 & 64.29 & \textbf{96.50} & \textbf{97.96} & \textbf{23.94} \\
  
    \bottomrule
  \end{tabular}
  }
  \label{tab:human_feedback}
\end{table}

\subsubsection{Impact of Progressive Iterative Training}

\label{secc:impact_of_progress}

\begin{wrapfigure}{r}{0.5\textwidth}
    \centering
    \includegraphics[width=\linewidth]{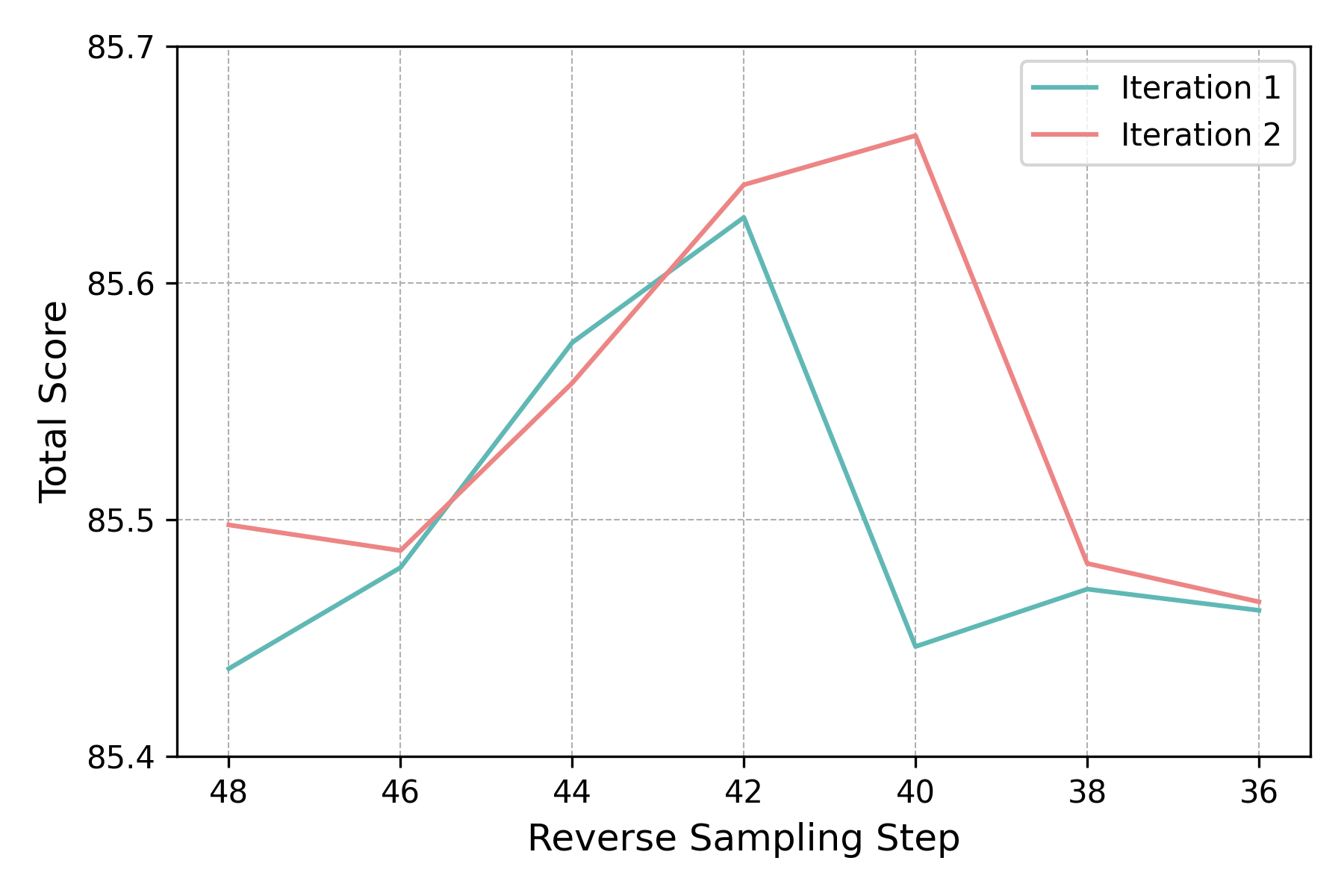}
    \caption{
    Total score across different reverse steps in training iterations 1 and 2.
    }
    \label{fig:final_score}
\end{wrapfigure}

To quantify the effect of our coarse-to-fine curriculum, we first plot RDPO performance versus the reverse-sampling step for consecutive training rounds; the curves are shown in Figure~\ref{fig:final_score}.  
Two clear trends emerge: (i) Reverse-step length matters. Within the first round (\textit{iter1}), using 42 reverse steps noticeably outperforms shorter trajectories, validating the need for rich, high-noise guidance at the beginning of training. (ii)Later rounds require fewer reverse steps. The \textit{iter2} curve is shifted to the right relative to \textit{iter1}, indicating that the model distribution, after one round of RDPO, has moved closer to the real-data manifold. Consequently, a smaller number of reverse steps is sufficient in subsequent rounds, improving training efficiency without sacrificing quality.

Furthermore, we monitor RDPO across three stages—\textit{iter1}, \textit{iter2}, and \textit{iter3}.  
The underlying LTX diffusion model employs 50 forward-diffusion steps, while RDPO follows a progressive schedule of reverse steps: 42 in \textit{iter1}, 40 in \textit{iter2}, and a mixture of 42- and 40-step data in \textit{iter3}.  
Table~\ref{tab:iter_training_effect} reports detailed metrics.  
Physics IQ improves monotonically over the three stages, and the total VBench score either rises or remains at a high plateau, confirming the effectiveness of the progressive curriculum.

\begin{table}[H]
  \caption{Effect of Iterative Progressive Training}
  \centering
  \renewcommand\theadfont{\normalsize\bfseries}
  \resizebox{\textwidth}{!}{ 
  \begin{tabular}{l *{13}{c}}
    \toprule
    \textbf{Models} 
    & \makecell{Physics\\IQ} 
    & \makecell{Total\\Score} 
    & \makecell{I2V\\Score} 
    & \makecell{Qulity\\Score} 
    & \makecell{Subject\\Consistency} 
    & \makecell{Background\\Consistency} 
    & \makecell{Motion\\Smoothness} 
    & \makecell{Aesthetic\\Quality} 
    & \makecell{Imaging\\Quality} 
    & \makecell{I2V\\Subject} 
    & \makecell{I2V\\Background} 
    & \makecell{Camera\\Motion} \\
    \midrule
    LTX-Video  & 24.40  & 85.35 & 92.85 & 77.85 & 96.76 & 98.36 & 99.25 & 54.62 & 63.96 & 96.33 & 97.83 & 22.41 \\
    RDPO (iter1)  & 25.21 & 85.63 & 92.98 & 78.27 & \textbf{97.04} & \textbf{98.81} & 99.27  & 55.01 & 65.11 & 96.39 & 97.87 & 23.72 \\ 
    RDPO (iter2)  & 25.31 & 85.66 & 93.07 & 78.25 & 96.83 & 98.69 & 99.28  & 55.01 & 65.41 & 96.42 & 97.87 & 25.24 \\
    RDPO (iter3) & \textbf{25.66} & \textbf{85.89} &\textbf{ 93.13} & \textbf{78.66} & 96.91 & 98.74 & \textbf{99.28}  & \textbf{55.14} & \textbf{65.71} & \textbf{96.44} & \textbf{97.91} & \textbf{25.69} \\
    
    \bottomrule
  \end{tabular}
  }
  \label{tab:iter_training_effect}
\end{table}




\subsection{Experiments on the \texorpdfstring{Shopee-MUG-V‑10B}{Shopee-MUG-V‑10B} Baseline}

\subsubsection{Implementation Details}



Given the larger parameters of Shopee-MUG-V-10B, we target more challenging generation tasks involving intricate human interactions and everyday scenes, aiming to improve physical consistency in these settings. 

To effectively train and evaluate the Shopee-MUG-V-10B model on these challenging tasks requiring enhanced physical consistency in intricate human interactions and everyday scenes, we curated a dedicated dataset. This dataset, tentatively named the Daily Physics Interactions Dataset (DPI), comprises videos specifically selected for their rich depictions of diverse physical phenomena occurring within realistic everyday settings and involving complex human-object or human-environment interactions. Examples include detailed sequences of object manipulation, collisions, fluid spills, soft body deformations, and complex movements that highlight underlying physical principles. The dataset was primarily sourced from online video collections. To ensure a robust evaluation of the model's generalization capabilities, we strictly held out a non-overlapping 10\% of the total videos to constitute the test split, with the remaining 90\% designated for training and validation.

Measuring physical consistency is challenging for existing automatic metrics, thus, we primarily rely on human evaluation for this crucial aspect. Human evaluators assessed videos based on four key criteria: \textit{Overall Preference}, \textit{Visual Quality}, \textit{Motion Naturalness}, and \textit{Physical Consistency}. For a more comprehensive assessment and reference, we also report standard automatic video quality metrics, including Subject Consistency, Background Consistency, Motion Smoothness, and Quality Score.



\begin{figure*}[!htbp]
    \centering
    \includegraphics[width=\textwidth]{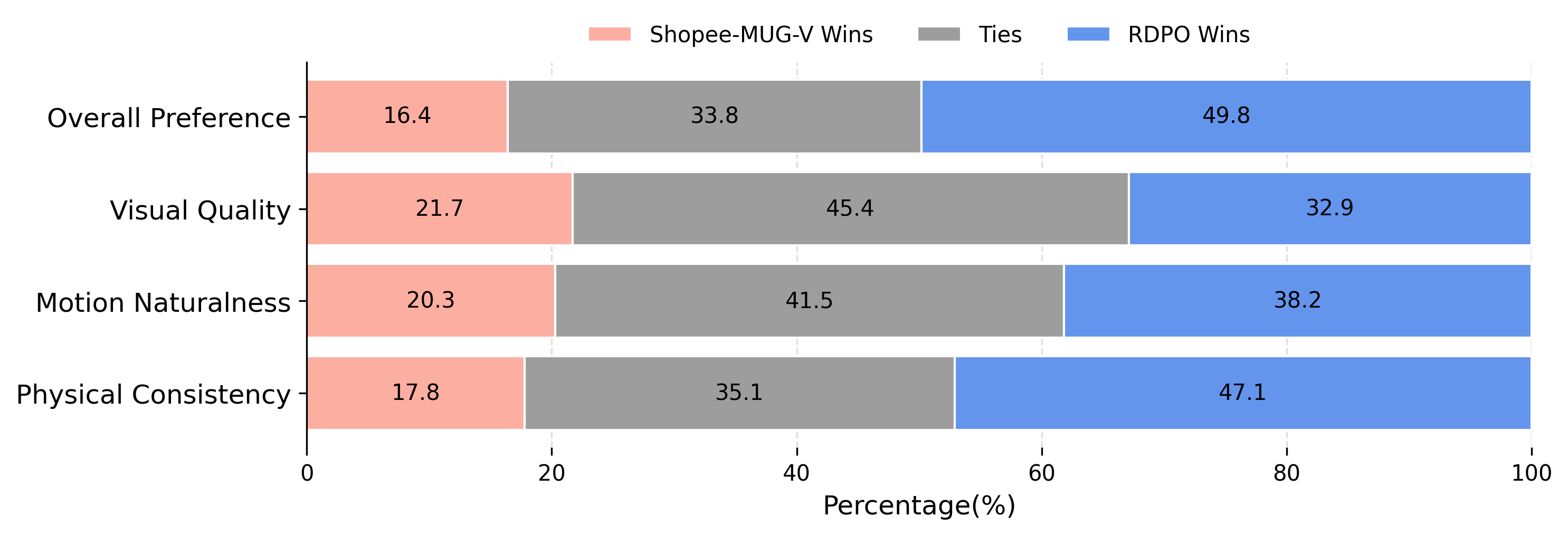}
    \caption{
    Human evaluation of RDPO and Shopee-MUG-V model.
    }
    \label{fig:huamn_evaluation}
\end{figure*}

\subsubsection{Experimental Results}

This section presents the experimental results for the Shopee-MUG-V-10B baseline, evaluating the performance of the base model compared to the model fine-tuned with our RDPO approach.

The results of our human evaluation study, summarized in Figure~\ref{fig:huamn_evaluation}, indicate that the Shopee-MUG-V model fine-tuned with RDPO was consistently preferred in pairwise comparisons across all four evaluated dimensions: Overall Preference, Visual Quality, Motion Naturalness, and Physical Consistency. The most pronounced improvements relative to the baseline were observed in Physical Consistency.

\begin{figure*}[!htbp]
    \centering
    \includegraphics[width=\textwidth]{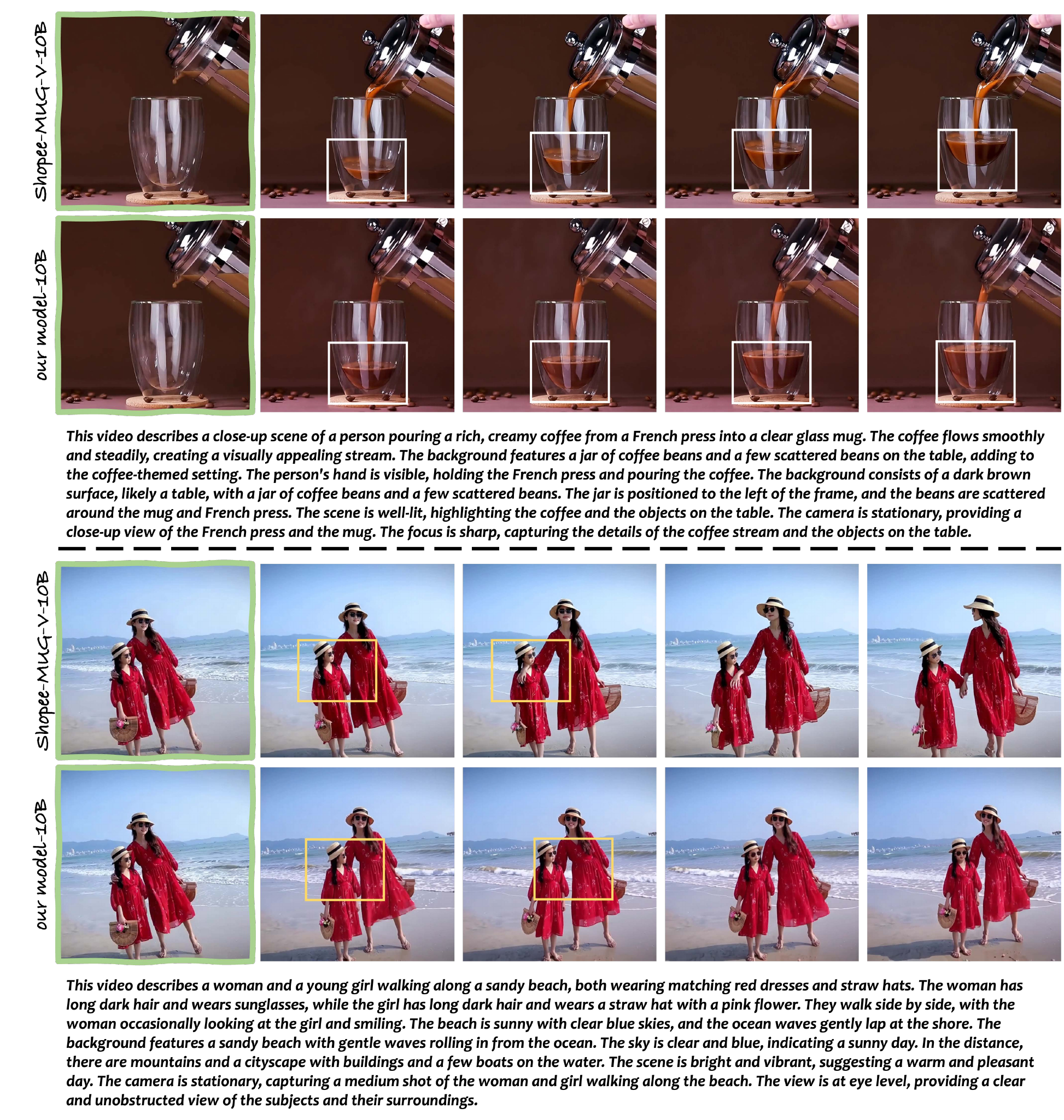} 
    \caption{
    Qualitative comparison between the Shopee-MUG-V model and our RDPO model. As shown in the figure, our method significantly improves the physical consistency, motion realism of the generated results. These qualitative improvements demonstrate the effectiveness of our approach.
    }
    \label{fig:comp_with_baseline}
\end{figure*}

Automatic metric results, presented in Table~\ref{tab:s1_auto_metrics}, likewise corroborate these findings from the human study. Quantitative comparisons show that the Shopee-MUG-V-RDPO model outperforms the base Shopee-MUG-V-10B model across all reported automatic video quality metrics: Subject Consistency, Background Consistency, Motion Smoothness, and Imaging Quality.

\begin{table}[H]
  \caption{Comparison of RDPO with other open-source model}
  \centering
  \renewcommand\theadfont{\normalsize\bfseries}
  \resizebox{\textwidth}{!}{ 
   \begin{tabular}{lcccccc}
    \toprule
    Model   &  Subject Consistency & Background Consistency & Motion Smoothness  & Imaging Quality  \\
    \midrule
    Shopee-MUG-V  & 96.48 & 97.59 & 99.25 & 75.45 \\
    Shopee-MUG-V-RDPO   &  97.41 & 97.98 &  99.61  & 76.37  \\
    \bottomrule
  \end{tabular}
  }
  \label{tab:s1_auto_metrics}
\end{table}

Qualitative examples showcasing the capabilities of the Shopee-MUG-V-RDPO model compared to the baseline are provided in Figure~\ref{fig:comp_with_baseline}. These visual comparisons further support the quantitative and human evaluation results, demonstrating notable improvements in physical consistency and motion realism achieved by our approach. Additional videos can be found in the supplementary material. Collectively, these results demonstrate that applying the RDPO approach effectively enhances the performance of the Shopee-MUG-V-10B model on challenging generation tasks, leading to improved physical consistency and overall video quality.

\section{Conclusion}

We introduced Real Data Preference Optimization (RDPO), an innovative annotation-free framework that significantly enhances the physical consistency of generated videos by directly leveraging physical priors from real-world video data. Through automated preference pair construction, RDPO circumvents costly manual annotations. Our experiments demonstrate that RDPO substantially improves physical realism and overall video quality across different model scales. Notably, RDPO achieves performance comparable to human-annotated DPO with significantly less effort and shows synergistic benefits when combined. RDPO offers a promising and scalable path towards generating videos with greater physical fidelity, reducing reliance on manual supervision.

\newpage

\appendix

\section{Dissusion}

\subsection{Statistical Validation of RDPO's Annotation-Free Pairwise Data}

\begin{wrapfigure}{r}{0.5\textwidth}
    \centering
    \includegraphics[width=\linewidth]{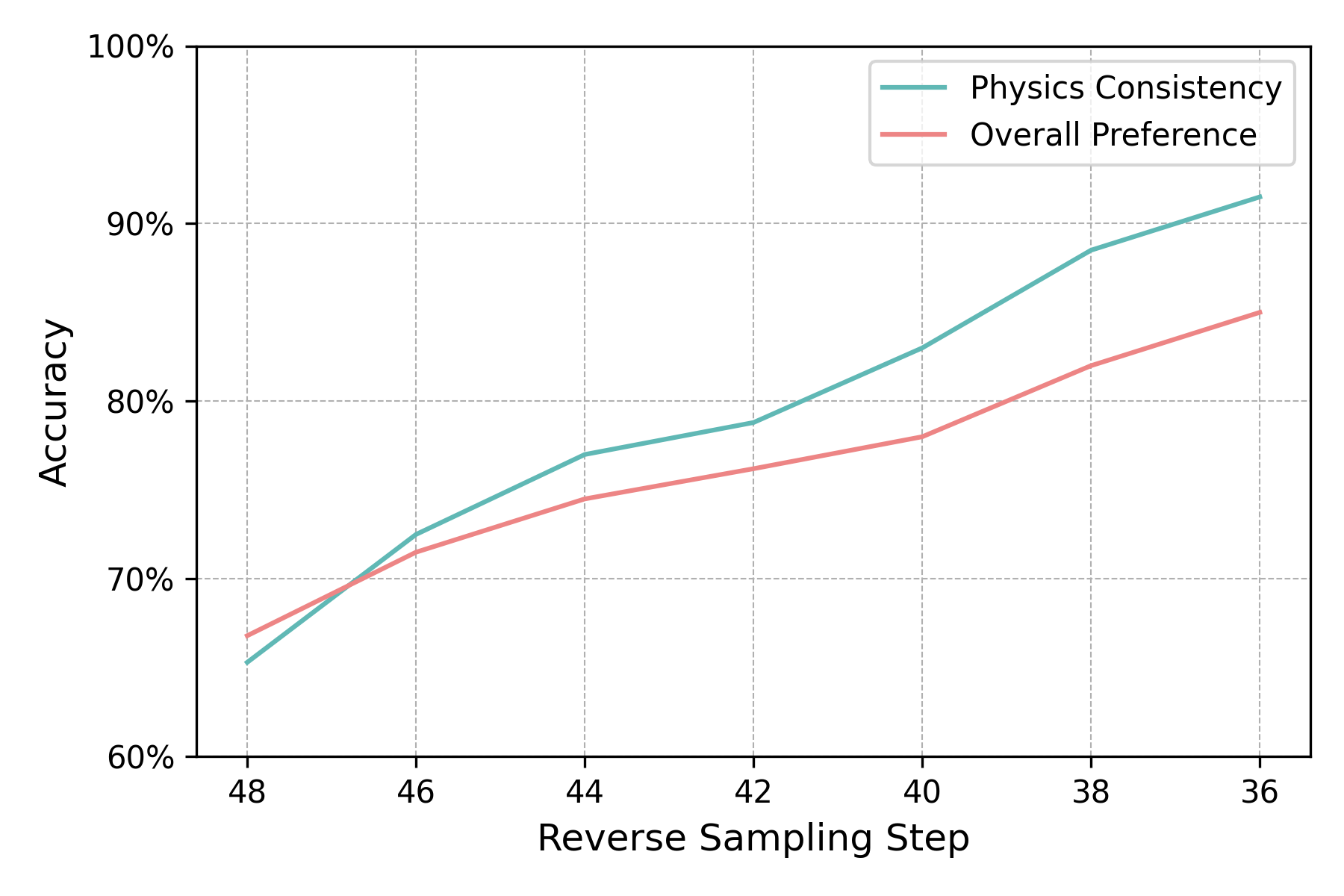}
    \caption{
    Accuracy of RDPO-Constructed Preference Pairs Under Human Evaluation. The graph demonstrates increasing alignment between RDPO's automated pair labeling and human judgments of physical consistency as the "positive" samples are derived using fewer reverse sampling steps from real-world video latents.
    }
    \label{fig:pairwise_accuracy}
\end{wrapfigure}

Real Data Preference Optimization (RDPO) leverages physical priors from real videos to automatically generate preference pairs, consisting of positive and negative samples. To demonstrate the effectiveness of this annotation-free paired data generation strategy, we conducted a dedicated human evaluation study.

In this study, human annotators were presented with video pairs automatically generated by RDPO. After a brief introduction on how to identify physical authenticity, their task was to select the video from each pair that exhibited greater physical consistency.

As illustrated in Figure \ref{fig:pairwise_accuracy}, the accuracy of the pairwise data constructed by RDPO, when compared to human judgment, shows a clear trend: accuracy improves as the number of reverse sampling steps used to generate the "positive" sample decreases. This result statistically confirms our assertion that RDPO's annotation-free approach can effectively construct positive and negative preference data pairs that align with human perception of physical realism.

Further analysis of the data highlighted that the distinctness of the preference was linked to the difference in reverse sampling steps between the paired videos. Specifically, when the "positive" sample was generated with considerably fewer reverse sampling steps , human annotators were more likely to agree with RDP's implicit labeling. This indicates that when a "positive" sample is generated closer to the latent distribution of real-world data , its superior physical plausibility becomes more evident to human observers. This increased clarity in distinction allows for the construction of more robust and reliable preference pairs. Ultimately, this validation process confirms that RDPO’s strategy effectively generates useful preference data for enhancing physical realism, all without the need for manual human annotation in the primary training loop.

\subsection{More Visual Qualitative Comparison}

\begin{figure*}[!htbp]
    \centering
    \includegraphics[width=\textwidth]{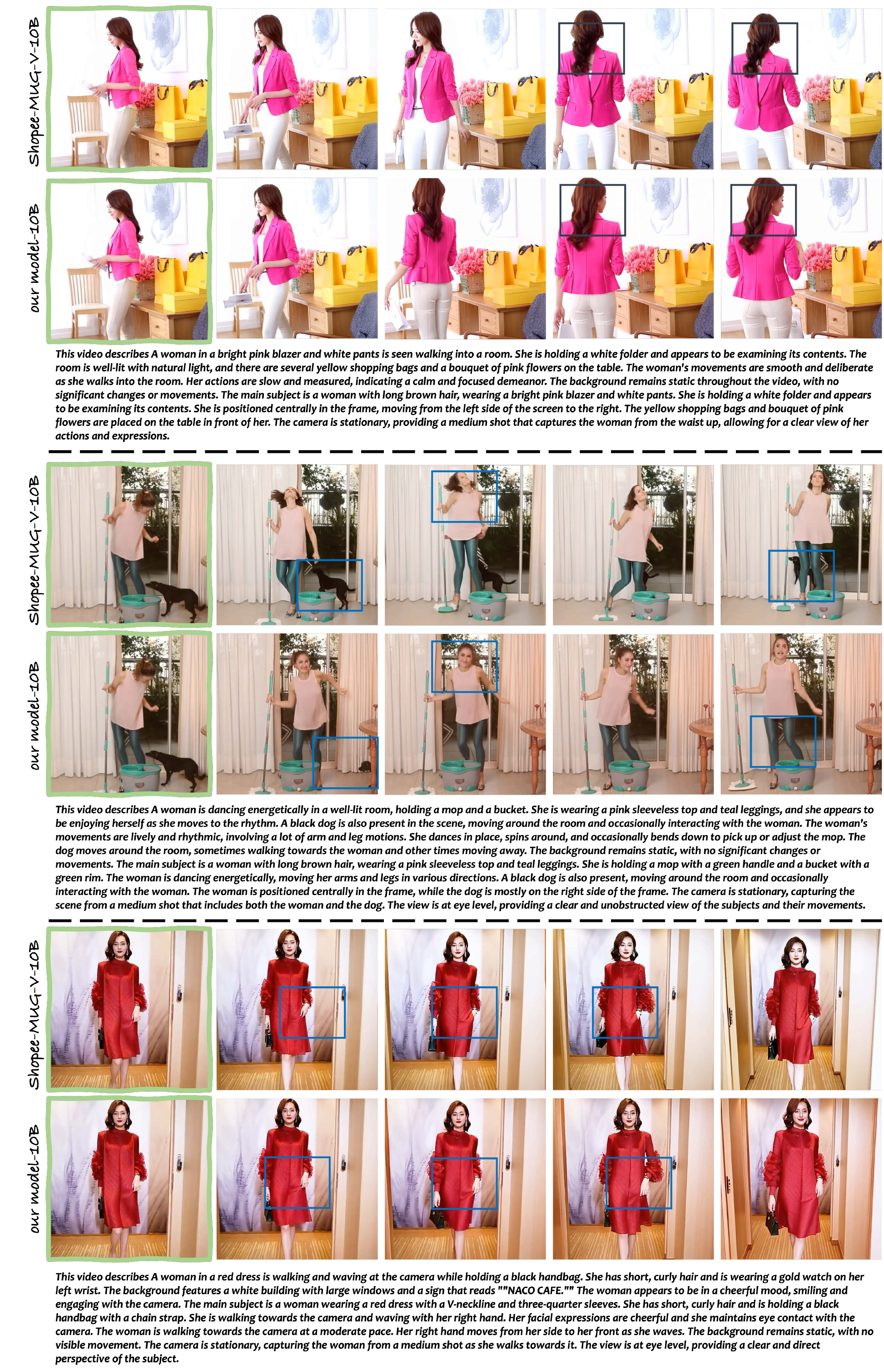}
    \caption{Qualitative comparisons showcasing the impact of RDPO on the Shopee-MUG-V. RDPO improves human kinematic coherence, such as coordinated head-body movements during turns. Furthermore, it enhances the naturalness of complex actions like dancing, while ensuring objects adhere to environmental constraints and character actions maintain contextual plausibility.}

    \label{fig:supply_img_1}
\end{figure*}

\begin{figure*}[!htbp]
    \centering
    \includegraphics[width=\textwidth]{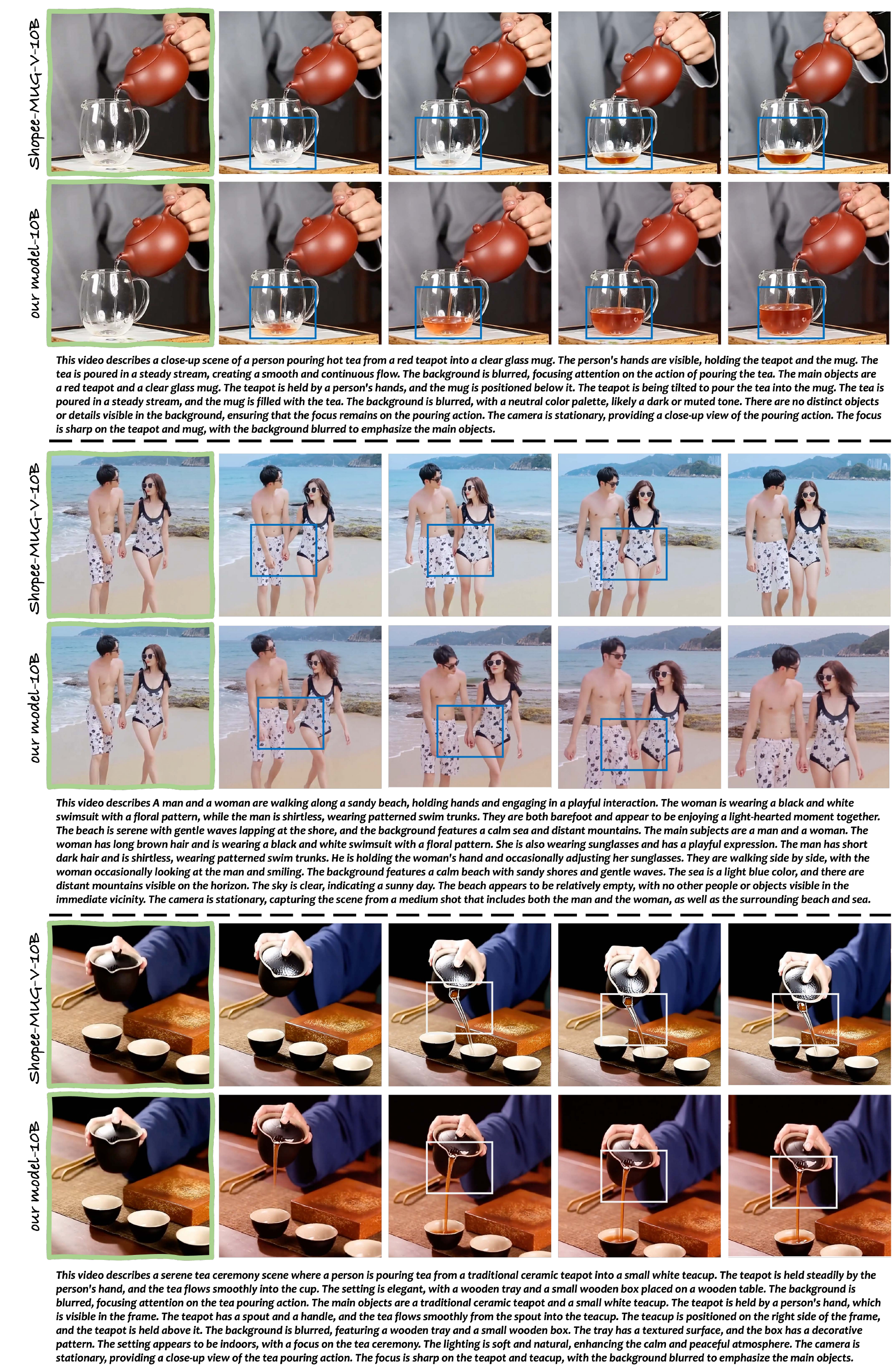}
    \caption{Qualitative comparisons showcasing the impact of RDPO on the Shopee-MUG-V, focusing on fluid dynamics and interactions. RDPO enables more accurate simulation of liquid accumulation during pouring and realistic fluid streams under gravity. It also improves the temporal stability of multi-character interactions, such as maintaining consistent arm positioning during hand-holding.}
    \label{fig:supply_img_2}
\end{figure*}

\begin{figure*}[!htbp]
    \centering
    \includegraphics[width=\textwidth]{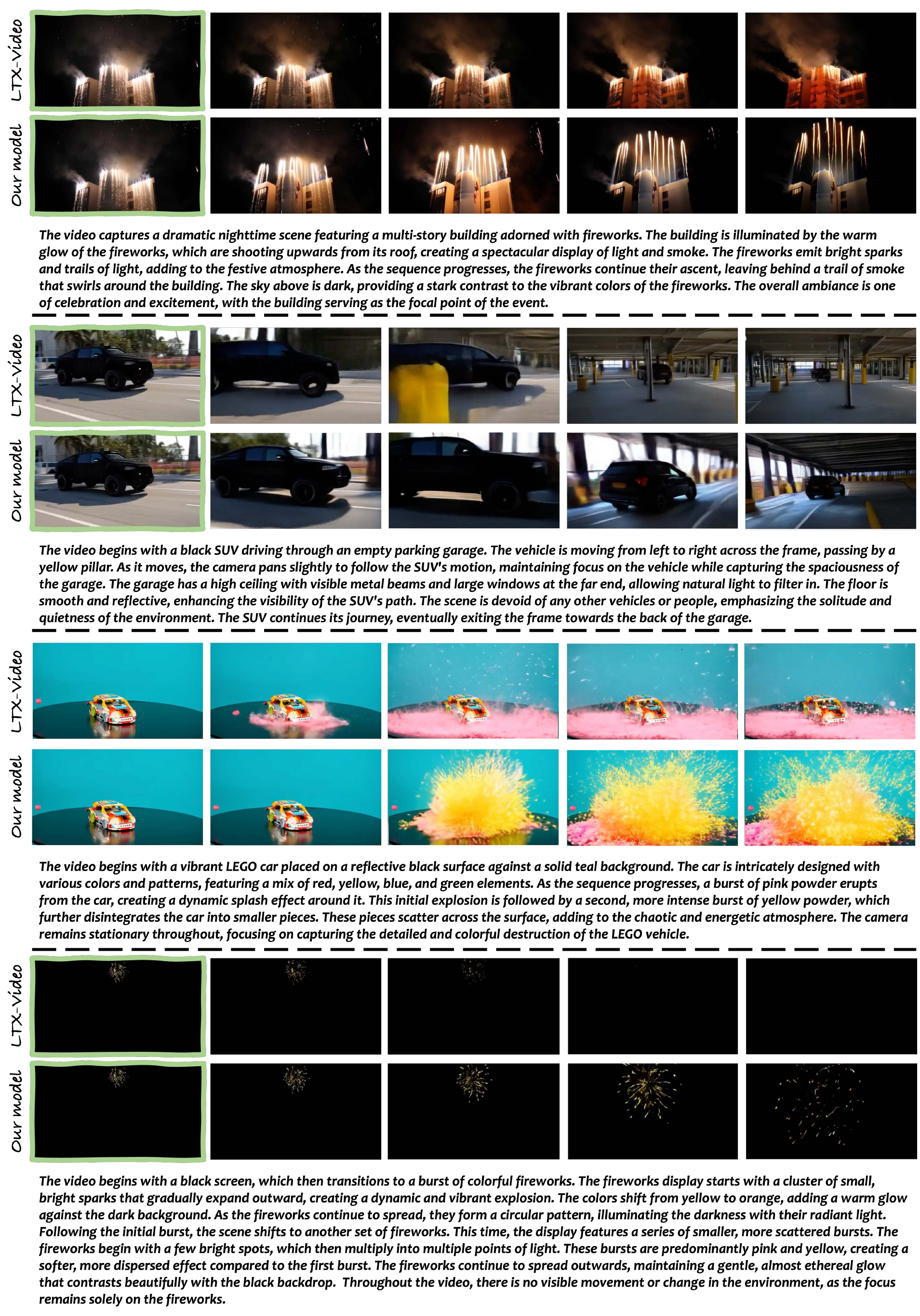}
    \caption{Qualitative comparisons showcasing the impact of RDPO on the LTX model, highlighting improvements in dynamic effects and scene coherence. RDPO facilitates more realistic depiction of particle-based effects like fireworks and enhances the naturalness of camera movements following dynamic subjects. Additionally, it contributes to the visual stability and contextual appropriateness of scene elements.}
    \label{fig:supply_img_3}
\end{figure*}

\begin{figure*}[!htbp]
    \centering
    \includegraphics[width=\textwidth]{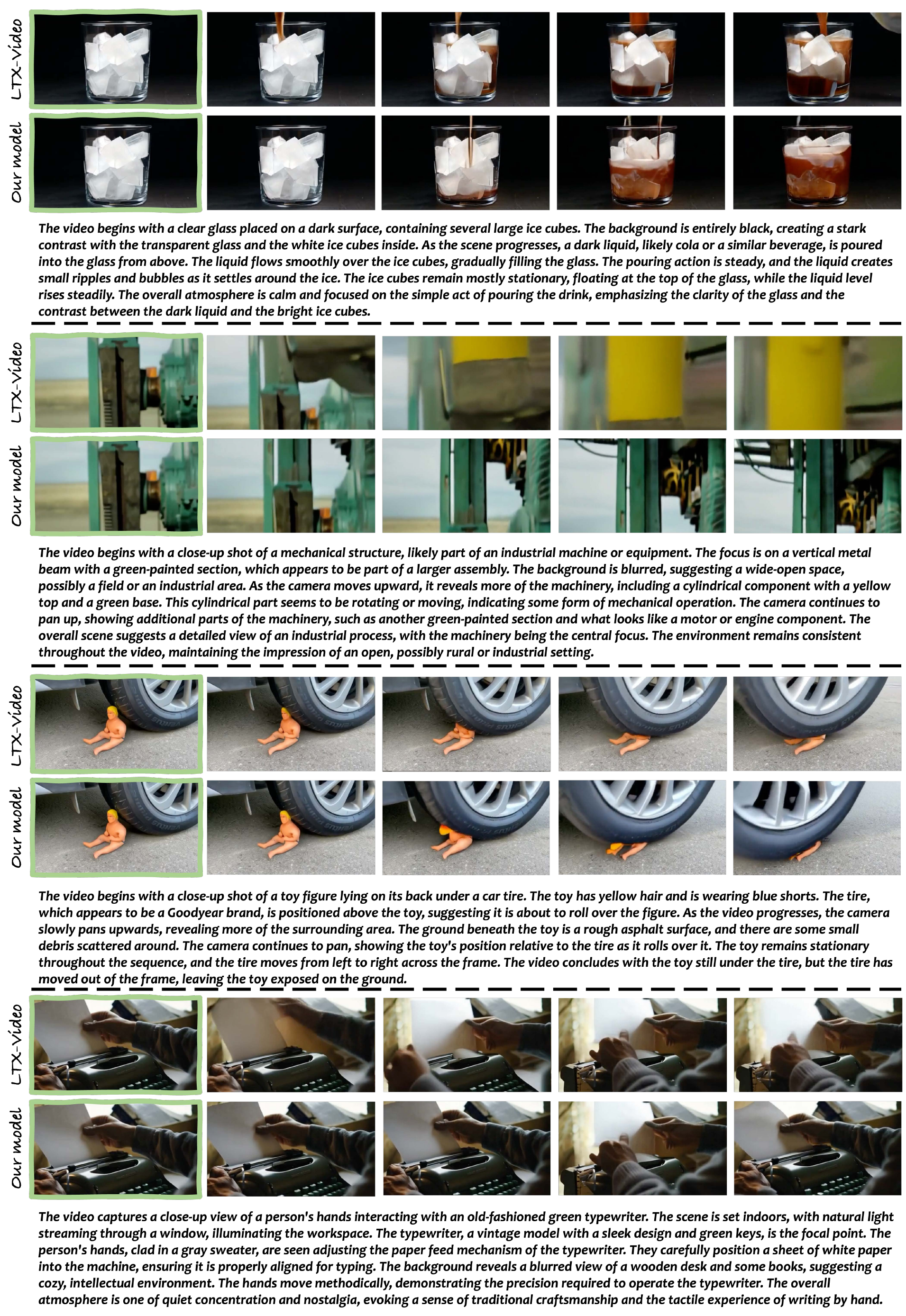}
    \caption{Qualitative comparisons showcasing the impact of RDPO on the LTX model, demonstrating RDPO's impact on complex physical interactions. Improvements include realistic multi-phase fluid behaviors, coherent rendering of intricate machinery, plausible deformable body dynamics under stress, and more nuanced human-object manipulation.}
    \label{fig:supply_img_4}
\end{figure*}

To qualitatively demonstrate the efficacy of RDPO as a versatile enhancement method, we present a series of comparative video examples. The illustrations in Figures~\ref{fig:supply_img_1}, \ref{fig:supply_img_2}, \ref{fig:supply_img_3}, and \ref{fig:supply_img_4} showcase improvements across multiple dimensions, including physical consistency and motion coherence, when comparing RDPO-enhanced models against their respective baselines. Specifically, we demonstrate advancements in the following areas: enhanced fidelity of human-object interactions and environmental physics; improved adherence to kinematic plausibility for both human and object motion; and heightened overall temporal consistency and realism in the depiction of dynamic events.

Across a broad range of evaluation scenarios, the original baseline models, Shopee-MUG-V and LTX, despite their commendable performance, often exhibit certain characteristic limitations prior to the application of RDPO. These limitations typically include: kinematic trajectories that are not entirely natural, occasional incoherence or instability within motion sequences, object-environment interactions that sometimes deviate from physical realism, and insufficient detail in the simulation of specific physical phenomena. Such issues can diminish the overall perceived realism of the generated videos.

When compared against their respective original baselines, models enhanced by RDPO consistently exhibit significant improvements. The RDPO framework markedly enhances the physical realism across diverse scenes and actions. Key improvements encompass: character animations that are smoother, more natural, and kinematically sound; object manipulations and interactions that more closely reflect real-world physical behaviors; more verisimilar simulations of fluid and deformable body dynamics; and notably enhanced causal consistency and continuity of physical events over time within the overall scene. These enhancements are evident regardless of the specific underlying architecture of the Shopee-MUG-V or LTX baselines, underscoring RDPO's robustness and general applicability in elevating the physical realism of generated video content.

\section{Limitations and Future Work}

While Real Data Preference Optimization (RDPO) demonstrates notable success in enhancing the physical consistency of generated videos, several limitations should be acknowledged, which concurrently outline promising directions for future research. Firstly, our current experiments have predominantly utilized datasets curated with common physical phenomena. To expand RDPO's capacity to model a broader and more diverse range of physical laws, the collection and curation of more varied real-world video data, specifically encompassing these phenomena, will be necessary. The effectiveness of RDPO is intrinsically linked to the quality and representativeness of the "real data" serving as the source of physical priors.

Secondly, as our training datasets were primarily selected for their depiction of physical regularities, this study has concentrated on improving physical consistency and, consequently, the overall perceptual quality of the videos. 
The current RDPO framework has not yet been explicitly optimized or evaluated for enhancing other distinct video attributes, such as narrative coherence, emotional expressiveness, or stylistic consistency across extended sequences. Future investigations could adapt the RDPO methodology by employing datasets specifically constructed to exemplify these alternative attributes, thereby exploring its potential to improve a wider spectrum of qualitative dimensions within the field of video generation.

Lastly, owing to constraints in computational resources and time, our empirical validation of RDPO was conducted primarily on models with 2B and 10B. Its performance on significantly larger or more advanced state-of-the-art foundation models remains an open area for investigation. We anticipate that as baseline video generation models continue to improve, their outputs will achieve increasingly high levels of visual fidelity. In such scenarios, discerning subtle physical inaccuracies may become considerably more challenging for human annotators. Therefore, RDPO, by directly leveraging real-world physical priors to generate preference pairs, could offer a more sensitive and objective optimization signal than traditional human preference-based methods. Consequently, as the capabilities of underlying models advance, the utility and potential superiority of RDPO are expected to increase. Future research will attempt validation on advanced models.

\newpage
\printbibliography{}




\end{document}